\documentclass{article}
\usepackage{ijcai26}

\usepackage{times}
\usepackage{soul}
\usepackage{url}
\usepackage[hidelinks]{hyperref}
\usepackage[utf8]{inputenc}
\usepackage[small]{caption}
\usepackage{graphicx}
\usepackage{amsmath}
\usepackage{amsthm}
\usepackage{amssymb}
\usepackage{booktabs}
\usepackage{algorithm}
\usepackage{algorithmic}
\usepackage[switch]{lineno}
\usepackage{multirow}
\usepackage{enumitem}
\setlist[itemize]{noitemsep}

\newtheorem{postulate}{Postulate}
\newtheorem{definition}{Definition}

\title{Towards Defining, Measuring and Leveraging Creativity in AI Applications}

\author{
Jingyi Yang
\and
Alexander Tuzhilin\\
\affiliations
New York University, Stern School of Business\\
\emails
\{jy4057, at2\}@stern.nyu.edu
}

\begin{document}

\maketitle

\begin{abstract}
Despite extensive research, computational creativity still lacks a conceptually grounded, practically relevant, and empirically applicable formal definition across contexts such as art, scientific research, and games. Existing approaches to creativity often focus on proposing informal conceptual definitions without providing formal computational concepts. Alternatively, they propose formal definitions without carefully examining their underlying assumptions or providing validation of these definitions by testing their performance in specific application settings. To address these problems, this paper proposes several formal postulates of creativity grounded in prior psychological work on creativity, especially as related to Boden’s three core criteria of creativity: newness, value, and surprise. Furthermore, we propose a conceptual mathematical definition of creativity that aligns with these postulates. We also validate our proposed creativity formula in two applications: determining the creativity of particular chess moves, and assessing the creativity of paintings by some artists, such as Picasso, Renoir, and other painters. Experimental results demonstrate that our formula of creativity successfully captures creative patterns across both the chess and the art domains. This suggests that creativity can be effectively characterized through a unified mathematical structure based on three fundamental criteria, namely newness, value, and surprise, whose specific measurements are adapted to the particular domains.
\end{abstract}

\section{Introduction}

The concept of creativity is crucial in many applications, including art, music, poetry, scholarly research, and advertising. Therefore, it has been studied for a long time across various disciplines, including psychology, philosophy, art, and computer science. Some scholars define creativity only conceptually by identifying the key criteria of creativity (e.g. novelty and value), without providing any formal mathematical definitions \cite{runco2012standard,stein1953creativity}, while others have tried to define it formally, including attempts to produce creativity formulas \cite{Pease2001,franceschelli2022deepcreativity}. One problem with this work is that very few proposed approaches have tried to carefully examine the underlying assumptions of creativity and also empirically validate their methods on real data to determine how accurately these methods measure creativity. 

To address these problems, we propose several postulates of creativity that are grounded in prior psychological work, especially as related to Boden's work \cite{boden2004creative}, where she identifies three core criteria of creativity: newness, value, and surprise. In other words, for an object or a concept to be creative, it should be new, have certain value, and be surprising \cite{boden2004creative}. Furthermore, we propose a mathematical definition of creativity that satisfies these creativity postulates.

We also validate our proposed creativity formula across two applications: determining the creativity of particular moves of chess pieces in the game of chess and assessing the creativity of paintings by some famous artists \footnote{Code, dataset descriptions, and supplementary materials accompanying this paper are publicly available at \url{https://github.com/MissTiny/IJCAI2026-Measuring-Creativity}.}. Note that the formula of creativity remains the same across these two applications; however, the specific definitions of newness, value, and surprise vary somewhat across the domains, making the ultimate measure of creativity application-dependent. For example, a creative move in chess is evaluated differently from the creativity of a painting, although both applications depend on the newness, value, and surprise criteria. 

In this paper, we make the following contributions. We
\begin{itemize}[topsep=0pt]
\item introduce several formal postulates of creativity that are based on prior research in psychology;
\item introduce a mathematical formula of creativity; 
\item test the proposed creativity formula on two applications, chess and paintings, and demonstrate that our definition of creativity works well on these two applications.
\end{itemize}

\section{Background and Related Work}\label{background}

The concept of creativity has been (informally) studied for centuries, mainly by philosophers and psychologists \cite{kaufman2016creativity}. A major shift occurred in 1950 when \cite{guilford1950creativity} proposed that creativity should be studied scientifically as a formal research field. Since then, various authors have proposed numerous approaches to understanding and capturing creativity without reaching a clear consensus on what creativity is \cite{treffinger1995creativity}. 

In particular, \cite{rhodes1961analysis} proposed that creativity can be studied from the \textit{``4P's of Creativity''} perspective: \textit{Person}, \textit{Process}, \textit{Products}, and \textit{Press}. The \textit{Person} perspective focuses on personal creativity and examines factors that make a person creative, while \textit{Process} perspective examines how creative concepts are generated as part of creative learning. The \textit{Product} perspective examines the factors that make an object creative, where objects refer to pieces of music, artwork, poems, etc. This perspective attracts the most attention within the creativity studies. Finally, the \textit{Press} perspective focuses on relationships between humans and their environments, including their culture(s). Moreover, Boden also proposes the \textit{Historical} vs. \textit{Psychological} dichotomy of creativity \cite{boden2004creative}. Historical creativity (\textit{H-creativity}) refers to the originality of a concept throughout history, while psychological creativity (\textit{P-creativity}) refers to the case when something appears to be new only to a person who created it. In this paper, we focus on the \textit{Product} creativity by providing  a formal definition and proper justification of what makes objects, such as paintings and chess games, creative from both the historical and personal perspectives.

Numerous approaches to defining Product creativity have been proposed in the literature. The classical two-criterion view holds that creativity requires only \textit{originality} and \textit{effectiveness} \cite{runco2012standard,stein1953creativity}. \textit{Originality} requires a product to be both \textit{unusual} and \textit{unique} where uniqueness is the primary emphasis. \textit{Effectiveness}, often equated with value, is inherently domain-dependent: for an artwork, value may be reflected in its market price or measured through quantitative aesthetic evaluations by human raters. In contrast, the highly influential and widely adopted three-criterion framework proposed by Boden \cite{boden2004creative} argues that a creative product must also be \textit{surprising}, interpreted as having unusual or non-obvious aspects that are not reducible to originality. This position has attracted substantial scholarly support; most notably, Simonton \cite{simonton2012taking} drew on U.S. Patent Office criteria\footnote{\href{http://www.uspto.gov/inventors/patents.jsp}{(http://www.uspto.gov/inventors/patents.jsp)}} to support Boden's three-criterion framework.

In the field of computational creativity (CC), AI can be artificially creative if its outputs are considered creative when generated by humans \cite{wiggins2019framework}. As with the status of defining creativity in the psychological field, the criteria for being creative in computational creativity have not yet reached a uniform consensus. For example, \cite{ritchie2007some} argued that computational creativity has the following three dimensions: \textit{novelty}, \textit{quality}, \textit{typicality}. Novelty and quality are defined similarly to originality and effectiveness in the psychological field, whereas typicality measures how closely the produced item resembles the artifact class.

In the context of creativity score estimation, one common approach is to directly map object embeddings to human-labeled creativity scores. For example, \cite{cropley2022automated} trained a CNN-based model to evaluate outputs from the Creative Thinking–Drawing Production (TCT-DP) task. While fine-tuning a pre-trained MobileNet model with human-annotated scores is effective, the embedding-based approaches lack interpretability into the underlying structure of creativity.

Other common approaches to estimating creativity via multiple criteria (e.g., newness, value, and surprise) have been previously proposed in the literature. These approaches exhibit the following limitations:

\paragraph{(1) Lack of theoretical grounding.} Some approaches incorporate features without clear connections to established theories of creativity, leading to spurious or non-generalizable results. For instance, \cite{KarampiperisComputational} defines storytelling creativity with four aspects: novelty, surprise, rarity, and recreational effort, where not all components are grounded in theories of storytelling or creativity. In contrast, our framework follows Boden's three-criterion perspective \cite{boden2004creative}, ensuring a theoretical foundation.

\paragraph{(2) Counterintuitive formulations.} Certain formulations contradict widely held intuitions about creativity. For example, \cite{franceschelli2022deepcreativity} proposes an additive formulation ($C = N + V + S$), allowing an object to have a positive creativity score even when one essential component (e.g., newness) is zero, which is counter-intuitive and also violates our Postulate (\ref{var: CreativityP}) proposed in Section \ref{defandpos}. In contrast, our multiplicative formulation requires all three components.

\paragraph{(3) Lack of implementation.} 
Many papers discuss creativity conceptually without concrete implementation details. For example, \cite{simonton2012taking} suggests that creativity arises from being new, valuable, and surprising and has a multiplicative relationship among these criteria, requiring each of them to be positive. However, \cite{simonton2012taking} does not provide concrete measurements for each criterion of creativity or empirically validate the formula on real-world applications.

\paragraph{(4) Ambiguity in measurement.} Even when creativity criteria are defined, the methods measuring them are poorly justified. For example, \cite{francca2016regent} measures novelty using Bayesian surprise, conflating novelty with surprise and blurring the distinction between them.

In contrast, our work, starting with psychologically and philosophically grounded postulates, proposes a principled and measurable mathematical formulation of creativity based on newness, value, and surprise. We further provide concrete methodologies for measuring each criterion, and empirically validate the framework across two distinct domains.

\section{Definition of Creativity and Postulates}\label{defandpos}
In this section, we follow the widely recognized and well-established Boden's view that an object is creative if it is \textit{new}, \textit{surprising}, and \textit{valuable} \cite{boden2004creative} and mathematically define creativity in terms of these criteria as follows.

\begin{definition}
\label{var: DefinitionP}
Let $\mathbf{X}$ denote the space of all possible objects. Then, $\mathbf{X}^\mathcal{D}$ represents the set of all possible objects within the domain $\mathcal{D}$. Let $\mathbf{T} \subseteq \mathbf{X}^\mathcal{D}$ be the set of pre-existing objects known to the domain $\mathcal{D}$. The creativity of an object $x \in \mathbf{X}^\mathcal{D}$ based on pre-existing objects $\mathbf{T}$, is denoted as $C^\mathcal{D}_{\mathbf{T},x}$, and is defined as a function of three components:
\begin{itemize}
    \item \textbf{Newness} $N^\mathcal{D}_{\mathbf{T},x} \in \mathbb{R}_{0}^{+}$: the degree to which $x$ differs from existing objects in $\mathbf{T}$.
    \item \textbf{Value} $V^\mathcal{D}_{\mathbf{T},x}\in \mathbb{R}_{0}^{+}$: the utility of $x$ in the domain $\mathcal{D}$.
    \item \textbf{Surprise} $S^\mathcal{D}_{\mathbf{T},x}\in \mathbb{R}_{0}^{+}$: the extent to which $x$ is surprising given domain knowledge.
\end{itemize}

The \textbf{creativity} of object $x$ for domain $\mathcal{D}$ is given by:
\begin{align}
    C^{\mathcal{D}}_{\mathbf{T},x} = f\left(N^{\mathcal{D}}_{\mathbf{T},x}, V^{\mathcal{D}}_{\mathbf{T},x}, S^{\mathcal{D}}_{\mathbf{T},x} \mid \alpha_{\mathcal{D}}, \beta_{\mathcal{D}}, \gamma_{\mathcal{D}} \right) \in \mathbb{R}_{0}^{+} 
\end{align}

\noindent
where $f$ is a continuous and differentiable function, and $\alpha_{D}, \beta_{D}, \gamma_{D}$ represent domain-dependent importance of each component.

\end{definition}

Definition (\ref{var: DefinitionP}) incorporated several previously discussed concepts about creativity. First, creativity is widely recognized as domain-specific, meaning that the measurement and relative importance of \textit{newness}, \textit{surprise}, and \textit{value} can vary across domains and applications \cite{baer1993creativity}. Therefore, parameters $\alpha_{\mathcal{D}}, \beta_{\mathcal{D}}, \gamma_{\mathcal{D}}$ are introduced to capture domain-specific preferences. Second, as discussed earlier, creativity can be categorized into \textit{historical} (H-creativity) and \textit{psychological} (P-creativity) \cite{boden2004creative}. Our definition accommodates both H- and P-creativity by introducing the pre-existing object set $\mathbf{T}$ containing \emph{all historical} objects in case of H-creativity and \emph{all known personal} objects in case of P-creativity. Lastly, we maintain that creativity, newness, value, and surprise scores should all be non-negative real numbers, where zero means that such a characteristic does not exist in the object.

Next, we propose eight intuitive postulates to serve as the guidance for creativity formula design. 

\begin{postulate}[Boundedness]\label{var: BasicPBound}
The creativity score of an object $x$, $C^{\mathcal{D}}_{\mathbf{T},x}$, as well as its components: newness $N^{\mathcal{D}}_{\mathbf{T},x}$, value $V^{\mathcal{D}}_{\mathbf{T},x}$, and surprise $S^{\mathcal{D}}_{\mathbf{T},x}$ must be bounded. Let $M^{\mathcal{D}}_C$, $M^{\mathcal{D}}_N$, $M^{\mathcal{D}}_V$, $M^{\mathcal{D}}_S$ denote the maximum value of creativity, newness, value, and surprise, respectively. Mathematically, 
\begin{align}
M^{\mathcal D}_C &= \max_{x \in \mathbf X^{\mathcal D}, \forall T \subseteq \mathbf{X}^{\mathcal D}} C^{\mathcal D}_{\mathbf T,x} \\
M^{\mathcal D}_N &= \max_{x \in \mathbf X^{\mathcal D}, \forall T \subseteq \mathbf{X}^{\mathcal D}} N^{\mathcal D}_{\mathbf T,x} \\
M^{\mathcal D}_V &= \max_{x \in \mathbf X^{\mathcal D},\forall T \subseteq \mathbf{X}^{\mathcal D}} V^{\mathcal D}_{\mathbf T,x} \\
M^{\mathcal D}_S &= \max_{x \in \mathbf X^{\mathcal D}, \forall T \subseteq \mathbf{X}^{\mathcal D}} S^{\mathcal D}_{\mathbf T,x} 
\end{align}
\begin{equation}
\begin{aligned}
&\exists\, M^{\mathcal{D}}_C, M^{\mathcal{D}}_N, M^{\mathcal{D}}_V, M^{\mathcal{D}}_S \in \mathbb{R}, \text{s.t.} \forall x \in \mathbf{X}^{\mathcal{D}}, \forall T \subseteq \mathbf{X}^{\mathcal D}\\
& C^{\mathcal{D}}_{\mathbf{T},x} \le M^{\mathcal{D}}_C, N^{\mathcal{D}}_{\mathbf{T},x} \le M^{\mathcal{D}}_N, V^{\mathcal{D}}_{\mathbf{T},x}\le M^{\mathcal{D}}_V,  S^{\mathcal{D}}_{\mathbf{T},x} \le M^{\mathcal{D}}_S
\end{aligned}
\end{equation}
\end{postulate}

Creativity is considered constrained by prior assumptions and resources \cite {ward2002role,de2010science}. Since resources are structured by prior knowledge, it is unlikely that the evolution of cognitive creativity will be entirely unbounded \cite{quine1969ontological}. Therefore, Postulate (\ref{var: BasicPBound}) states that there exists an upperbound value for each of the four variables. For instance, if the value of an artwork is measured by auction price in the art domain, then $M^{\mathcal{D}}_V$ is naturally bounded by the total wealth available in that market.

\begin{postulate}[Continuity]\label{var: BasicPCont}
The creativity score $C^{\mathcal{D}}_{\mathbf{T},x}$ is determined by three variables: $N^{\mathcal{D}}_{\mathbf{T},x}$, $V^{\mathcal{D}}_{\mathbf{T},x}$,and $S^{\mathcal{D}}_{\mathbf{T},x}$, where creativity function $f$ is continuous over its entire domain.
\end{postulate}

As discussed in \cite{boden2004creative}, creativity should be a continuous numeric value (vs. a binary one). 

\begin{postulate}[Creativity Property]\label{var: CreativityP}
An object $x \in \mathbf{X}^{\mathcal{D}}$ is considered creative if and only if it exhibits all three essential attributes of creativity:
\begin{itemize}
    \item Newness: $N^{\mathcal{D}}_{\mathbf{T},x} > 0$
    \item Value: $V^{\mathcal{D}}_{\mathbf{T},x}> 0$
    \item Surprise: $S^{\mathcal{D}}_{\mathbf{T},x} > 0$
\end{itemize}

Formally,
\begin{align}
        C^{\mathcal{D}}_{\mathbf{T},x} > 0 \iff N^{\mathcal{D}}_{\mathbf{T},x} > 0,\ V^{\mathcal{D}}_{\mathbf{T},x} > 0,\ S^{\mathcal{D}}_{\mathbf{T},x} > 0
\end{align}
\end{postulate}

Postulate (\ref{var: CreativityP}) illustrates that an object is not creative if an object lacks \emph{any} of the three components. Considering a newly designed keyboard with a known structure but a different alphabet arrangement, this new keyboard is new and valuable, assuming someone wants to use it. However, the new keyboard is not surprising, as people can expect all possible rearrangements given the limited combination options. Therefore, the keyboard design is not creative. As discussed in Section \ref{background}, this postulate is violated by additive formulations of creativity. For example, \cite{franceschelli2022deepcreativity} defines creativity as the sum of newness, value, and surprise, allowing an object to be creative based on only its newness or value without necessarily exhibiting the other characteristics, which conflicts with Postulate (\ref{var: CreativityP}).

\begin{postulate}[Newness Property]\label{var: NewnessP}
Let $x' \in \mathbf{X}^{\mathcal{D}}$ be a new object and let $\mathbf{T} = \{x_1, x_2, \dots, x_n\} \subseteq \mathbf{X}$ be the set of known objects. The newness score $N^{\mathcal{D}}_{\mathbf{T},x'}$ reflects the level of difference between $x'$ and all $x_i \in \mathbf{T}$.
\begin{itemize}
    \item If $\forall  x_i \in \mathbf{T}, x' \neq x_i$ that is $x'$ is different from all objects in $\mathbf{T}$,  then $N^{\mathcal{D}}_{\mathbf{T},x'}> 0$.
    \item If $\exists  x_i\in \mathbf{T}, x' = x_i$ that is $x'$ is identical to  one of the objects in $\mathbf{T}$, then $N^{\mathcal{D}}_{\mathbf{T},x'} = 0$.
\end{itemize}
\end{postulate}
Postulate (\ref{var: NewnessP}) states that the newness of object $x$ means that it is different from all other objects in set $\mathbf{T}$. 
However, we have not yet defined what it means for an object to be identical to an existing object. This definition should be tailored to the specifics of the domain. For example, in the art domain, replicating Vincent van Gogh's ``The Starry Night'' on a T-shirt is considered equivalent to the original painting itself, since the content of the painting image remains unchanged regardless of the medium.

\begin{postulate}[Newness-Surprise Separability Property]\label{var: NewsupP}
Newness $N^{\mathcal{D}}_{\mathbf{T},x}$ and surprise $S^{\mathcal{D}}_{\mathbf{T},x}$ are separable properties of an object $x$; that is, neither implies the other.

\begin{align}
\exists x \in \mathbf{X}^{\mathcal{D}}, s.t. N_{\mathbf{T},x}> 0, S_{\mathbf{T},x} = 0\\
\exists x \in \mathbf{X}^{\mathcal{D}}, s.t. N^{\mathcal{D}}_{\mathbf{T},x} = 0, S^{\mathcal{D}}_{\mathbf{T},x} > 0\\
\exists x \in \mathbf{X}^{\mathcal{D}}, s.t. N^{\mathcal{D}}_{\mathbf{T},x} = 0, S^{\mathcal{D}}_{\mathbf{T},x} = 0\\
\exists x \in \mathbf{X}^{\mathcal{D}}, s.t. N^{\mathcal{D}}_{\mathbf{T},x} > 0, S^{\mathcal{D}}_{\mathbf{T},x} > 0
\end{align}
\end{postulate}

\begin{postulate}[Newness-Value Separability Property]\label{var: ValuenewP}
Newness $N^{\mathcal{D}}_{\mathbf{T},x}$ and value $V^{\mathcal{D}}_{\mathbf{T},x}$ of an object $x$ are separable; that is, neither implies the other.

\begin{align}
\exists x \in \mathbf{X}^{\mathcal{D}}, s.t. N^{\mathcal{D}}_{\mathbf{T},x}> 0, V^{\mathcal{D}}_{\mathbf{T},x} = 0\\
\exists x \in \mathbf{X}^{\mathcal{D}}, s.t. N^{\mathcal{D}}_{\mathbf{T},x} = 0, V^{\mathcal{D}}_{\mathbf{T},x} > 0\\
\exists x \in \mathbf{X}^{\mathcal{D}}, s.t. N^{\mathcal{D}}_{\mathbf{T},x} = 0, V^{\mathcal{D}}_{\mathbf{T},x} = 0\\
\exists x \in \mathbf{X}^{\mathcal{D}}, s.t. N^{\mathcal{D}}_{\mathbf{T},x} > 0, V^{\mathcal{D}}_{\mathbf{T},x} > 0
\end{align}
\end{postulate}

\begin{postulate}[Surprise-Value Separability Property]\label{var: ValuesupP}
Surprise $S^{\mathcal{D}}_{\mathbf{T},x}$ and value $V^{\mathcal{D}}_{\mathbf{T},x}$ of an object $x$ are separable; that is, neither implies the other.
\begin{align}
\exists x \in \mathbf{X}^{\mathcal{D}}, s.t. S^{\mathcal{D}}_{\mathbf{T},x}> 0, V^{\mathcal{D}}_{\mathbf{T},x} = 0\\
\exists x \in \mathbf{X}^{\mathcal{D}}, s.t. S^{\mathcal{D}}_{\mathbf{T},x} = 0, V^{\mathcal{D}}_{\mathbf{T},x} > 0\\
\exists x \in \mathbf{X}^{\mathcal{D}}, s.t. S^{\mathcal{D}}_{\mathbf{T},x}= 0, V^{\mathcal{D}}_{\mathbf{T},x} = 0\\
\exists x \in \mathbf{X}^{\mathcal{D}}, s.t. S^{\mathcal{D}}_{\mathbf{T},x}> 0, V^{\mathcal{D}}_{\mathbf{T},x} > 0
\end{align}
\end{postulate}

Separability property in Postulates (\ref{var: NewsupP}), (\ref{var: ValuenewP}), and (\ref{var: ValuesupP}) states that the newness, value, and surprise aspects are non-redundant: absence of one of them does not determine the presence or absence of the others. An object may be new but not valuable, and vice versa. For example, people can randomly draw lines on paper to produce a new painting; however, it may not be valuable at all. Similarly, people can reprint Van Gogh's paintings (these reprints are not new); but they can be valuable since people may buy them.

\begin{postulate}[Newness-Surprise Non-Continuity]\label{var: SurprisingNewnessP}
Given a prior dataset $\mathbf{T} \subseteq \mathbf{X}^{\mathcal{D}}$, the relationship between newness $N^{\mathcal{D}}_{\mathbf{T},x}$ and surprise $S^{\mathcal{D}}_{\mathbf{T},x}$ are discontinuous. That is, small variations in newness can correspond to large variations in surprise, and vice versa.
\end{postulate}

Newness and surprise are related concepts. Their relationship has been studied and debated previously by philosophers and psychologists. For example, \cite{wiggins2006preliminary} characterizes surprise as a response of the observer to a creative artifact, while \cite{francca2016regent} measures novelty using Bayesian Surprise–based methods without explicitly modeling surprise as a separate component. In contrast, \cite{macedo2003model,simonton2012taking} emphasize the importance of surprise as a separate aspect of creativity and make a strong case for its distinction from newness. 

In our work, we follow \cite{macedo2003model,simonton2012taking} and adopt the view that newness and surprise are two related but, nevertheless, distinct concepts. While Postulate (\ref{var: NewsupP}) states that the absence of newness does not determine the presence or absence of surprise (and vice versa), Postulate (\ref{var: SurprisingNewnessP}) captures a different and stronger relationship between these two aspects.  Specifically, large variations in newness may lead to small changes in surprise. For instance, the films \textit{The Magnificent Seven} and \textit{The Seven Samurai} differ substantially in their settings (American Western vs. feudal Japan). Yet, this large variation does not result in big surprise, as the underlying narrative structures are similar. This example illustrates that significant changes in newness (The Magnificent Seven is very different from The Seven Samurai) do not always translate into high surprise values. Similarly, small changes in newness can lead to large changes in surprise. For example, in the case of punchline jokes, a small change in wording can lead to a substantial change in surprise. The sentence,``To be frank, I’d have to change my personality'', is an ordinary statement and not particularly humorous. However, replacing a single word yields ``To be frank, I’d have to change my name,'' where the reinterpretation of \textit{frank} as a proper name introduces a surprising and humorous twist. Postulate (\ref{var: SurprisingNewnessP}) addresses that the relationship between newness and surprise can be discontinuous as discussed above, which is different from Postulate (\ref{var: NewsupP}).

\section{Creativity Formula}\label{measurement}
We propose to use the following closed-form creativity formula to mathematically measure the creativity score:
\begin{align}\label{eq:creativity}
    C^{\mathcal{D}}_{\mathbf{T},x} = {N^{\mathcal{D}}_{\mathbf{T},x}}^{\alpha_{\mathcal{D}}} {V^{\mathcal{D}}_{\mathbf{T},x}}^{\beta_{\mathcal{D}}} {S^{\mathcal{D}}_{\mathbf{T},x}}^{\gamma_{\mathcal{D}}} 
\end{align}
\noindent
where Equation (\ref{eq:creativity}) is a special case of a generic definition of creativity defined by Equation (\ref{var: DefinitionP}). Next, we provide particular ways to define newness (N), value (V), and surprise (S) measures used in Equation (\ref{eq:creativity}), which we do in the rest of this section.  

\subsection{Measure of Newness}
Since any measure of newness must satisfy Postulate (\ref{var: NewnessP}),
we propose the following definition of newness for object $x$:
\begin{align}\label{eq:newness}
   N^{\mathcal{D}}_{\mathbf{T},x} = \min_{x_i \in \mathbf{T}} dist(x,x_i)
\end{align}
where $dist(\cdot)$ represents the distance function, $\mathbf{T} \subseteq \mathbf{X}^\mathcal{D}$ is the set of pre-existing objects known to domain $\mathcal{D}$, and $x$ and $x_i \in \mathbf{T}$ are \textit{embedding} representations of the actual objects. Equation (\ref{eq:newness}) measures the minimum distance from the focal object to any objects in the existing set $\mathbf{T}$.

Multiple distance functions can be used to measure newness, such as the Euclidean or Cosine similarity distance. Furthermore, the minimum function is preferred over its alternatives, such as the median, average, or the distance to the center of the cluster, as the minimum gurantees Postulate (\ref{var: NewnessP}), whereas the median and average functions may not.

\subsection{Measure of Value}
We assume there is a way to measure an object's intrinsic value, or at least estimate it, using various methods. For example, in the case of artworks, the intrinsic value can be estimated by the maximum sale price of the art object at an auction. The maximum represents the highest willingness to pay for the artwork over an extended period, reflecting its long-term value, as is the case with Vincent van Gogh's paintings.

We can also generalize this argument by assuming that the ``business'' value V of object $x$ fluctuates over time, and we take this maximum business value over time as an intrinsic value of $x$:
\begin{align}\label{eq: valuable1}
    V^{\mathcal{D}}_{\mathbf{T},x} =  V^{\mathcal{D},I}_x \approx \max_t V^{\mathcal{D},E}_t
\end{align}
where $V^I$ represents the intrinsic value of object $x$ and $V^{\mathcal{D},E}_t$ represents its business value at time $t$ in the domain $\mathcal{D}$.

An intrinsic value can also be evaluated through domain-specific criteria as: 
\begin{align}\label{eq: valuable2}
    V^{\mathcal{D}}_{\mathbf{T},x} =  V^{\mathcal{D},I}_x \approx \sum_{r_i \in \mathbf{R}} r_{i,x}
\end{align}
where $\mathbf{R}$ is the set of criteria that can be used to measure the value of an object according to domain experts and $r_i$ is a criterion from $\mathbf{R}$. For example, coherence and interestingness can be used when measuring creativity of a writing \cite{hosanagar2025designinghumangenerativeai}. When evaluating creativity of a move in a chess game, the value can be estimated by the change in the winning rate after each move \cite{zaidi2024predicting}. 

In general, when intrinsic measures are available and reliable, they should be used. Otherwise, external values, such as market prices and user feedback, can be used as substitutes to approximate value.

\subsection{Measure of Surprise}\label{supmeasure}
The techniques used to measure surprise can be divided into three categories: \textit{probabilistic mismatch}, \textit{belief mismatch}, and \textit{observation mismatch}  \cite{MODIRSHANECHI2022102712,dinparastdjadid2023}. 
Although all three are useful, belief mismatch measures surprise by comparing the prior and posterior belief distributions before and after observing a new object. When the size of the prior knowledge is large, this distributional shift is negligible, making belief mismatch an insensitive measure for evaluating creativity. Therefore, the probabilistic and observation mismatch categories are more suitable for measuring surprise in creativity than the belief mismatch category, and we focus on them in this paper. 

In \textbf{probabilistic mismatch} surprise measures, surprise is defined as the likelihood of an object $x$ existence given the prior distribution. Shannon's Surprise \cite{Shannon1948} and Bayesian Factor Surprise \cite{Xu2021} are typical examples of this type of measure. For example, Shannon's Surprise is defined as the negative log probability of an object $x$ given prior probability distribution $P$ learned from the set of objects $\textbf{T}$. The prior distribution over $\textbf{T}$ can be modeled using Gaussian Mixture Models (GMMs), a probabilistic model that represents the objects in $\textbf{T}$ as mixtures of Gaussian distributions. 

\textbf{Observation mismatch} surprise measures the distance between the predicted object $\hat{x}$ and the actual observation $x$. The absolute error and square error between the \textit{embeddings} of the expected new object $\hat{x}$ and the real new object $x$ can be considered as the observation mismatch surprise. 

There are multiple ways to obtain $\hat{x}$. For example, $\hat{x}$ can be obtained through a generative model trained with $\mathbf{T}$, although training a model iteratively through updating with incoming data is computationally intensive. Therefore, in the chess domain, where the set of legal chessboard states is finite, a dictionary-based approach is feasible and $\hat{x}$ can be inferred from a dictionary constructed from prior knowledge. In contrast, for domains such as artwork and storytelling, where the set of possible future objects is infinite, dictionary-based methods are not suitable, and the probabilistic mismatch measures are more appropriate. 

In this section, we introduced the specific creativity formula and explained how to compute its three key components: newness, value, and surprise. Our discussion highlights that the measurement of each component must be carefully tailored to the characteristics and constraints of the application domain. In the next section, we apply the creativity formula to two use cases, paintings and chess, and show how well our method works in these two applications.

\section{Framework Validation on Two Applications}\label{application}
To validate our framework, we applied Equation (\ref{eq:creativity}) to the Chess Games and Painting Artworks applications described below.

\subsection{Chess Game}
\label{sec:chess}
Chess is considered an application related to the concept of creativity. For example, \cite{zaidi2024predicting} developed a method to predict the `brilliant" moves perceived by humans.  The authors view ``brilliance'' in chess as closely related to the concept of creativity, though they admit that ``brilliant'' and ``creative'' are not the same philosophically. According to chess.org's definition \cite{MasterMatthew52_2023_brilliant}, a ``brilliant'' move in chess requires an unexpected effect that has a significant impact. Furthermore, brilliant moves require a ``good piece sacrifice,'' according to chess.org, which differs from the definition of creativity discussed in this paper.

We evaluated creativity formula (\ref{eq:creativity}) for chess on the dataset described below, as explained in this subsection.

\paragraph{Dataset.} Within the Chess application, we measure the creativity score of a chess move within a game. We use the dataset Mega ChessBase 2025, which contains chess games played between 1610 and 2024 \cite{megachessbase2025} and some of the chess games are annotated by the chess experts. Comments include opening strategy, move analysis, tactical implications, creativity assessments, and the endgame strategy. Of particular relevance are the comments explicitly characterizing creativity of a move. The dataset fields contain the names of two players, their ranks, moves, and the game results. After cleaning the dataset by removing the chess games with incomplete or faulty data, the revised dataset contains 94,565 games with 7,711,327 chess moves in total.

\paragraph{Object Representation.} The chess board position represented by Forsyth-Edwards Notation (FEN), a standard notation for describing positions in a chess game with only text strings, is converted into embeddings using the pretrained model ChessLM \cite{hull2025beyond}. The embedding of a chess move is computed as the difference between the FEN embedding of the board state immediately after the move and that immediately before it. All the chess games and their corresponding moves that took place before 1960 form the initial prior knowledge set $\textbf{T}$ in our definition of creativity. Overall, we have 10,842 games in set $\textbf{T}$, all played before 1960. The prior knowledge set $\textbf{T}$ is updated yearly. For example, to evaluate the creativity score of a chess move played in 2021, all moves played before 2021 are considered as prior knowledge. Then, when evaluating the chess move in 2022, chess games played in 2021 will be added to the prior knowledge set $\textbf{T}$, along with games before 2021.

\paragraph{Newness.} A chess move is ``new'' (equal to 1) if it has never occurred before in any other game in the collection of the previously played games, and is ``old'' (equal to 0) otherwise. In particular, we adopt a binary definition of newness in chess because, given the finite nature of the board and the presence of well-established openings and game strategies, it is highly likely that a focal move has already been played by many experts under the same board positions. Therefore, representing newness as a binary variable (0 for previously observed, 1 for unseen) is sufficiently informative, as truly novel moves are rare in large historical game databases. Moreover, small variations between moves, while potentially meaningful from a strategic perspective, often carry limited additional information in terms of novelty. Instead, the frequency of a move conditioned on the board position provides a more informative signal, which is captured by our surprise component.

\paragraph{Value.} Stockfish \cite{CCRL_40_15_All_Engines} is a popular publicly available chess engine. As a part of its algorithm, Stockfish computes a value of its moves and makes this computation publicly available. Furthermore, this value is positive if the White is more likely to win the game and negative if the Black is more likely to win. Therefore, we directly used this value, as computed by the Stockfish engine with a depth of 15 steps ahead, in this study. 

\paragraph{Surprise.} We used the observation mismatch as a surprise measure in chess and computed it as follows. Given a preexisting set of games $\textbf{T}$, we constructed a lookup dictionary in which each key is a board position occurring in $\textbf{T}$, and each corresponding value is itself a dictionary mapping every move played at that position to the number of times it was played across all prior games. The surprise value of a move is computed as the distance between the move's embedding and the weighted average of the embeddings of all the preexisting moves in the current board position taken from the inner-dictionary,  

\paragraph{Evaluation.} To evaluate our approach, we derived ground truth creativity labels from the expert-annotated comments provided in the dataset. Specifically, any move described by an expert using terms such as ``creative'' or semantically equivalent language defined by Merriam-Webster Dictionary was a potential candidate for a creative move (17,000 comments in total). To ensure a high-quality label, we further used three LLMs (`chatgpt 4o-mini', `gemini-2.5-flash-lite ', and `claude-3-5-haiku-20241022') that take expert comments about a particular chess move as input and determine if it is really related to creativity assessment or not (i.e., the expert may comment on other aspects unrelated to creative move). We prompted LLMs by asking the following questions: 1) What is the comment on? 2) Is creativity (novelty + value + surprise) explicitly mentioned or is it implied?  3) Is (are) the move(s) described as surprising? The LLM's answers to these questions are in the multiple choice form, e.g., (a) Positive, (b) Negative, (c) Not Related, (d) Not Sure. In addition to the multiple choice answers, a short justification paragraph is required for each question. Furthermore, robustness was ensured through consistency criteria: the pairwise cosine distance between the CLIP embeddings of justifications produced by three LLMs must be greater than 0.70, and all three LLMs return the same result (e.g., all three values are ``Positive''). As a result, running expert-based commentaries through LLMs and computing creativity-based metrics as described above can serve as a reasonable proxy for estimating perceived creativity. 

After applying the above creativity labeling procedure, we obtain 639 samples of chess moves labeled as being creative vs. non-creative. Motivated by the discrete and finite nature of the chess action space, the ground-truth creativity labeling choice is binary. Nevertheless, the creativity score remains a real value (according to Postulate (\ref{var: BasicPCont})), and is defined here as a continuous function predicting \textit{probability} of a move being creative.
The resulting dataset of 639 moves is imbalanced, with 78\% of the samples labeled as creative. Note, however, that this imbalance is not an artifact of the annotation process, as creative moves should be rare. Instead, it reflects the process of how experts comment chess games: they tend to comment predominately interesting and non-obvious chess moves, creativity being one of major categories of the moves on which they focus, thus attracting disproportionately large number of comments. As a result, when we filter the data to retain only comments that explicitly discuss creativity, the resulting dataset naturally over-represents creative moves, leading to a skewed 78\% distribution. 

\paragraph{Model.} When the outcome variable is binary, the parameters $\alpha_{\mathcal{D}}$,$\beta_{\mathcal{D}}$, and $\gamma_{\mathcal{D}}$ in Equation (\ref{eq:creativity}) can be learned through logistic regression by taking logarithms of both sides in Equation (\ref{eq:creativity}). We call this method the `basic model' in the paper. To capture potential non-linear relationships among variables, we applied the SplineTransformer \cite{perperoglou2019review,eilers1996flexible}, which expands each numerical feature into multiple B-spline basis functions, and enables linear models to fit non-linear patterns. This variant is referred to as the `non-linear model'.

\subsection{Painting Artworks}
The second creativity application in which we tested our creativity formula was Painting Artworks. Note that this application is much more complex than chess since the total space of all possible paintings is infinite.

\paragraph{Dataset.} We have acquired the fine-art auctions data reported by an anonymous source as of September 2nd, 2025. After extensive cleaning of this data, removing various types of dubious records, focusing only on the western art painters producing their work between the years 500AD and 2024, we ended up with 355,551 auction records having full information about the painting itself, the artist, the auction price in USD, and the verified date when the painting was made by the artist. Among them, 2,802 paintings have multiple auction records with a maximum of 4 auctions.

\paragraph{Object Representation.} The image of each painting is represented with an \textit{embedding} computed using classical computer vision methods. In order to test the robustness of our method, we have tried multiple popular pre-trained embeddings for Artwork Painting, including CLIP \cite{radford2021learning}, DINOv2 \cite{oquab2023dinov2}, 
and ViT \cite{deng2009imagenet} embeddings. 

\paragraph{Newness.} Unlike the binary Chess case, the newness of a painting is computed as a real-value through the function defined as the minimum distance between the embedding of a painting object and that of objects in the prior knowledge group $\mathbf{T}$ as shown in Equation (\ref{eq:artworknew}), which is a particular case of Equation (\ref{eq:newness}).
\begin{align}\label{eq:artworknew}
    N^{artwork}_{\mathbf{T},x} =  \min_{x_i \in \mathbf{T}} \| x - x_i \|_2
\end{align}
where $x$ is the embedding (e.g. CLIP embeddings) of the focal painting and $x_i$ is the embedding of a painting in set $\mathbf{T}$. This distance is $0$ if painting $x$ already belongs to group $\mathbf{T}$, which is unlikely in most of the real-world settings.

\paragraph{Value.} Estimating the true value of a painting is a very complex and subjective task \cite{coslor2016transparency}. To avoid this highly subjective activity, we use a proxy: estimating the value of a painting based on its auction price. Clearly, this is only one of many possible estimates of a painting's true value, since the auction price reflects only the highest bidder's willingness to pay for the artwork. Nevertheless, we maintain that this is a reasonable proxy since bidders at the high-stakes auctions are usually avid art collectors who pay a lot of money for the top paintings and can therefore rationally estimate the ``true'' value of a painting. As discussed before, the maximum auction price recorded for the artwork was used in this paper as the artwork's value. In addition, the values were adjusted to the year 2024 using the Consumer Price Index (CPI) to account for inflation and changes in the purchasing power of USD.

\paragraph{Surprise.} Unlike the observation mismatch surprise used in Chess, it is computationally much harder to do the same thing for the Artwork case. Therefore, we used the probabilistic mismatch surprise method in this case, including the Shannon Surprise measure discussed in Section \ref{supmeasure} that is defined as the negative log probability of object $x$ (e.g., artwork), given the pre-existing set of objects (prior artwork) $\textbf{T}$. Formally, a GMM is fitted on the embeddings of objects in $\mathbf{T}$ to approximate its underlying distribution, so that the likelihood of a new artwork $x$ appearing under this distribution can be estimated. A low likelihood indicates that $x$ deviates substantially from prior paintings and is thus more surprising.

\paragraph{Evaluation.} Unlike the chess dataset that contains human expert annotations on the creativity of chess moves, there is no expert criticism on artwork in the dataset. Therefore, we use LLMs with the search engine for labeling to avoid hallucination. We used `chatgpt 4o-mini', `gemini-2.5-flash-lite',`claude-3-5-haiku-20241022' with prompts that ask for comments on artwork creativity based on search results. Due to the limitation of the budget, we decided to focus only on the artworks from the following artists: Käthe Kollwitz, Pierre-Auguste Renoir and Pablo Picasso. The aforementioned LLMs produced comments on how creative particular paintings were. Only the comments with pairwise cosine distance computed based on the CLIP embeddings of creativity criticism produced by the three LLMs that are greater than 0.70 are considered consistent and valid. Since artworks are generally considered an area closely related to creativity, the LLM responses conclude that most of the artworks are creative. Therefore, we decided to make the ground-truth creativity label being numeric to measure the \emph{degree} of creativity of a painting. Furthermore, the Aspect-Based Sentiment Analysis (ABSA) model \cite{YangL24LCF/LCA,YangCL23PyABSA} is applied to compute the creativity score defined as the probability of being positively creative. Then each comment is evaluated using the ABSA model alongside `creativ' (note that with `creativ', the semantics of the similar words, such as `creativity' and `creative', will be analyzed). After filtering the paintings using the above criteria, we ended up with 1,862 labeled samples.

\paragraph{Model.} After normalizing the newness, value, surprise, and creativity score, we learn $\alpha_{\mathcal{D}}$, $\beta_{\mathcal{D}}$ and $\gamma_{\mathcal{D}}$ coefficients in Equation (\ref{eq:creativity}) by taking logarithms of both sides of the equation and converting it to the linear regression model:
\begin{align}
\label{eqn:creativity:log}
\log C^{\mathcal{D}}_{\mathbf{T},x} = &\alpha_{\mathcal{D}}\log N^{\mathcal{D}}_{\mathbf{T},x} +
\beta_{\mathcal{D}}\log V^{\mathcal{D}}_{\mathbf{T},x} \notag\\
&+ \gamma_{\mathcal{D}}\log S^{\mathcal{D}}_{\mathbf{T},x}
\end{align}
As in the Chess case, the above model serves as the `basic model' for the artwork domain. To introduce non-linearity, the SplineTransformer \cite{perperoglou2019review,eilers1996flexible} with degree 4 and 8 knots is applied, yielding the `non-linear model'.

\section{Results}\label{result}

\subsection{Chess Games}

Table \ref{tab:ChessResult} presents the performance of our creativity formula across multiple metrics, including accuracy, AUC, precision, recall, and F1. The results are reported as average performances based on the 5-fold cross-validation with the standard deviations shown right next to the averages.  

\begin{table}[]
\centering
{\fontsize{9pt}{11pt}\selectfont
\setlength{\tabcolsep}{4pt}
\begin{tabular}{lccccc}
\toprule

                 & Acc      & AUC           & Prec.     & Rec.        & F1            \\
\midrule
Basic  & .570\scriptsize$\pm$.06 & .618\scriptsize$\pm$.08 & .824\scriptsize$\pm$.04 & .571\scriptsize$\pm$.07 & .673\scriptsize$\pm$.05 \\
Non-linear & \textbf{.640\scriptsize$\pm$.04} & \textbf{.636\scriptsize$\pm$.07} & \textbf{.827\scriptsize$\pm$.02} & \textbf{.683\scriptsize$\pm$.05} & \textbf{.747\scriptsize$\pm$.03}\\

\bottomrule
\end{tabular}}
\caption{Chess Game Results}\label{tab:ChessResult}
\end{table}

Since the dataset is unbalanced, the AUC metric is our primary focus. Note that the AUC of 0.6357 (achieved by the non-linear model) is a satisfactory result for such a complex task as estimating creativity in chess. We sought to identify appropriate baselines for comparing our creativity formula in the chess application. Despite our extensive efforts, we could not identify any appropriate baselines that can be \textit{directly} comparable to our model. Most research on chess applications focused on making the correct move rather than a creative one \cite{granter2017alphago}. Although \cite{zaidi2024predicting} touches the subject of creativity a bit, its main focus is on predicting brilliant chess moves, which is distinct from creativity in chess, as explained earlier. The framework presented in \cite{fenggenerating} focuses on generating creative chess puzzles, and creativity evaluation (determining the ground truth) is performed by humans rather than a specific metric, which is not comparable with our work. 

Therefore, we compared the performance of our creativity estimation task with that of a task of similar difficulty. Specifically, \cite{werneburg2024external} reported that a proposed neural network for predicting clinical responses in a medical application achieved the AUC performance result of 0.60, which \cite{werneburg2024external} claimed to be a strong result allowing the proposed method outperform human experts. In comparison, our creativity formula's AUC performance is even higher than that in the clinical prediction task, despite both problems being very difficult.

To confirm the Separability Property of Postulates (\ref{var: NewsupP}), (\ref{var: ValuenewP}), and (\ref{var: ValuesupP}), correlation levels between newness and surprise, surprise and value, and newness and value are computed for chess, and they are equal to $0.4785$, $-0.0290$, $-0.1513$, respectively. These values indicate that newness and surprise are correlated to some extent, but are not collinear and therefore constitute distinct concepts that cannot be derived from one another.

\subsection{Painting Artworks}
Table \ref{tab:ArtnetResult} presents the performance of our creativity formula for the Artworks case across different types of embeddings (rows) and multiple metrics (columns), including MAE, RMSE, JSD and WD. Similar to the case of Chess, the results reported in Table \ref{tab:ArtnetResult} are the average performance numbers for the 5-fold cross-validation, and the standard deviations are shown next to them.

Since the label in our painting dataset represents the probability that a painting is creative, as estimated by the ABAS model, our model is trained to predict this probability. Therefore, in addition to standard regression metrics, such as RMSE and MAE, we also use the following two distribution evaluation metrics: 
\begin{itemize}
    \item \textbf{Jensen-Shannon Divergence (JSD)} (a.k.a \textit{Information Radius (IRad)}): JSD measures similarity between two probability distributions, i.e., the total divergence to the average, where a value close to zero indicates that the distribution of the true creativity label and the distribution of the predicted creativity are highly similar.
    \item \textbf{Wasserstein Distance (WD)}: A similarity metric between the distribution of true creativity labels and that of predicted creativity scores, measuring the minimum amount of probability mass that must be transported to transform one distribution into the other.
\end{itemize}

Among the three embeddings, the basic model performs better on MAE, RMSE, and JSD, whereas the non-linear model performs better on WD. Particularly, with the CLIP embedding, the basic model performs somewhat better with the 4\% in MAE, 9\% in RMSE and 20\% improvements in JSD vs. the non-linear model. Meanwhile, the non-linear model performs better with the 10\% improvement in WD vs. the basic model, indicating that it is more accurate point-wise and better at mass allocation. Similarly, performance improvements are around the same levels for the ViT and DINOv2 embeddings according to Table \ref{tab:ArtnetResult}. Our model achieves a JSD close to zero, indicating strong alignment between the predicted and true creativity distributions, which demonstrates the effectiveness of our approach. Similar to the case of Chess, we could not find a comparable baseline for the arts application. Papers on artwork creativity focus on artwork generation \cite{mazzone2019art} or unsupervised measurement \cite{elgammal2015quantifying}, which is not applicable to our data. Specifically, \cite{elgammal2015quantifying} proposed a creativity measurement with value and novelty without a ground truth to compare with. 

As a reference, we evaluated our creativity estimation performance by comparing it with that of a task of similar or even lesser difficulty. Specifically,  \cite{Apell_niz_2024} proposed a divergence-based framework for synthetic tabular data validation, reporting the best JSD of 0.099 on the Adult dataset using a VAE-based generative model. In contrast,our model's JSD performance is 0.0016, which is significantly better than the \cite{Apell_niz_2024}'s best reported result, thus demonstrating good JSD-based performance of our method for the Painting application.

\begin{table}[]
{\fontsize{7pt}{10pt}\selectfont
\setlength{\tabcolsep}{3pt}
\begin{tabular}{llcccc}
\toprule
Embedding               & Model            & MAE & RMSE  & JSD & WD \\ 
\midrule
\multirow{2}{*}{CLIP}   & Basic      & \textbf{.0287\scriptsize$\pm$.002}  & \textbf{.0441\scriptsize$\pm$.005}  & \textbf{.0016\scriptsize$\pm$.000}   & .0249\scriptsize$\pm$.002 \\
                        & Non-linear &  .0300\scriptsize$\pm$.002   &  .0487\scriptsize$\pm$.007     &.0020\scriptsize$\pm$.001   &  \textbf{.0223\scriptsize$\pm$.002}  \\ \hline
\multirow{2}{*}{ViT} & Basic      &  \textbf{.0287\scriptsize$\pm$.002}   & \textbf{.0441\scriptsize$\pm$.005}      & \textbf{.0016\scriptsize$\pm$.000}    &  .0248\scriptsize$\pm$.002  \\
                        & Non-linear &  .0300\scriptsize$\pm$.001   &  .0489\scriptsize$\pm$.003    &  .0019\scriptsize$\pm$.000   &  \textbf{.0225\scriptsize$\pm$.003}  \\ \hline
\multirow{2}{*}{DINOv2}    & Basic      & \textbf{.0287\scriptsize$\pm$.002}    & \textbf{.0441\scriptsize$\pm$.005} &  \textbf{.0016\scriptsize$\pm$.000 } & .0249\scriptsize$\pm$.002 \\
                        & Non-linear &  .0302\scriptsize$\pm$.003   &    .0498\scriptsize$\pm$.009  & .0020\scriptsize$\pm$.001   & \textbf{.0228\scriptsize$\pm$.003}\\
\bottomrule
\end{tabular}}
\caption{Painting Artwork Results}\label{tab:ArtnetResult}
\end{table}

To confirm the Separability Property in Postulates (\ref{var: NewsupP}), (\ref{var: ValuenewP}), and (\ref{var: ValuesupP}), we compute correlation levels between newness and surprise, surprise and value, and newness and value, which are equal to $0.2986$, $0.0308$, $0.0421$ respectively in the painting application based on the CLIP embedding. These values suggest that newness and surprise are correlated to some extent but are not collinear, confirming that newness and surprise capture distinct aspects of creativity and cannot be derived from one another.

\section{Conclusion and Future Plan}\label{conclusion}

In this paper, we formally define the concept of creativity in terms of its newness, value, and surprise criteria and show that it satisfies several postulates of creativity that are grounded in prior work in psychology. We also validate our proposed creativity formula on two applications: determining creativity of particular moves of chess pieces in the game of chess and assessing creativity of paintings by some artists. Although the creativity formula of  (\ref{eq:creativity}) remains the same across these two applications, the specific definitions of newness, value, and surprise vary across the domains, making the ultimate measure of creativity application-dependent. 

Future work includes applying our method to additional domains, such as storytelling, music and advertising, along with developing domain-specific solutions for each new application. Also, the current creativity formula represents one instantiation within the broader postulate framework, and we intend to investigate whether its canonical form exists.

\bibliographystyle{named}
\bibliography{ijcai26}

\end{document}